# ChevOpt: Continuous-time State Estimation by Chebyshev Polynomial Optimization

Maoran Zhu, *Student Member, IEEE*, Yuanxin Wu, *Senior Member, IEEE*

*Abstract*—In this paper, a new framework for continuous-time maximum a posteriori estimation based on the Chebyshev polynomial optimization (ChevOpt) is proposed, which transforms the nonlinear continuous-time state estimation into a problem of constant parameter optimization. Specifically, the time-varying system state is represented by a Chebyshev polynomial and the unknown Chebyshev coefficients are optimized by minimizing the weighted sum of the prior, dynamics and measurements. The proposed ChevOpt is an optimal continuous-time estimation in the least squares sense and needs a batch processing. A recursive sliding-window version is proposed as well to meet the requirement of real-time applications. Comparing with the well-known Gaussian filters, the ChevOpt better resolves the nonlinearities in both dynamics and measurements. Numerical results of demonstrative examples show that the proposed ChevOpt achieves remarkably improved accuracy over the extended/unscented Kalman filters and extended batch/fixed-lag smoother, closes to the Cramer-Rao lower bound.

## I. INTRODUCTION

The state estimation is to estimate the state of a system from the stochastic dynamic model and measurements. Thanks to many applications in spacecraft attitude determination [1, 2], robotics navigation [3, 4], biological process estimation [5], and many others, the state estimation algorithms have gained intensive research interests in the past decades.

The Bayesian framework, which computes the posterior density function of the state, provides an optimal solution to the state estimation [6]. For the linear and additive Gaussian noise system, the Kalman filter [7] is the optimal estimation method. However, for more general nonlinear systems with non-Gaussian noise, it is generally intractable to achieve the optimal estimation without any approximation.

Gaussian filters are a subset of Bayesian estimator for processing measurements up to the current time, under the assumption of Gaussian distribution for both states and noises [8-10]. A variety of Gaussian filters have been proposed in the literature to deal with the nonlinearity, which can be roughly classified into the Taylor expansion-based filter, the deterministic sampling-based filter and the Monte Carlo-based filter [10, 11]. The most celebrated filter is the extended Kalman filter (EKF) [12], which approximates the nonlinearity of the state dynamics and measurement functions by successive first-order Taylor expansion at current estimate. The EKF has been employed in vast practical applications, but it is prone to divergence due to the first order approximation. Second order Taylor expansion [13] is explored to decrease the approximation error, yet not commonly used in practice due to the cumbersome Hessian matrix. The iterated EKF is another strategy that tries to lessen the effect of the measurement nonlinearity by iteratively linearizing the measurement equation [13]. To further improve the state estimation accuracy, the backward smoother, which utilizes the measurements made before and after the current time, can be applied following the forward filter [14].

An alternative method to Gaussian filters for addressing nonlinearity is to utilize a set of deterministic sampling points to approximate the probability distribution. The unscented Kalman filter (UKF) [15, 16] generates the sigma points, according to the known prior density, and utilizes the unscented transformation to propagate statistical information of the posteriori density. The quadrature Kalman filter is proposed in [9, 10, 17] for both discrete and continuous systems from the standpoint of Gauss-Hermite numerical quadrature. The works [18, 19] utilize the cubature rule to numerically approximate integrals and build the cubature Kalman filter. The advantage of these sampling-based filters is that they avoid evaluating the cumbersome Jacobian/Hessian matrix and do not require dynamics and measurement functions to be smooth and differentiable. In contrast to the deterministic sampling filter, the Monte Carlo-based particle filter [20-22] draws a large number of random samples to approximate and propagate the probability distribution, which promises to be the best state estimation algorithm. However, the exponential increase of samples with the dimensionality of the state space leads to a great computation burden.

Another class of methods related to this paper is the optimization-based state estimation. The works [23, 24] propose an efficient incremental optimization algorithm based on the factor graph to decrease the computation burden in optimization-based batch estimation. And, [25-27] propose the so-called moving horizon estimation (MHE), which optimizes the discrete-time state in a sliding window with the objective function being formed by the dynamics, measurements and

This paper was supported in part by National Key R&D Program of China (2018YFB1305103).

Author's address: M. Zhu and Y. Wu are with Shanghai Key Laboratory of Navigation and Location-based Services, School of Electronic Information and Electrical Engineering, Shanghai Jiao Tong University, Shanghai 200240, China (email: zhumaoran@sjtu.edu.cn, yuanx_wu@hotmail.com).

assumed prior distribution. As for the continuous-time estimation, the works [28-30] initiate a batch maximum a posteriori (MAP) estimation method based on the B-spline function approximation for the simultaneous localization and mapping problem. However, the state estimate by the low-order B-spline is not accurate enough and the approximation error has to be remedied through the weighting matrix [31, 32]. Besides, the batch estimation cannot be applied to the real-time applications.

In this paper, we try to introduce the Chebyshev collocation method to solve the continuous-time state estimation problem. The collocation method is a well-known numerical algorithm for solving differential equations, which is to find a polynomial that satisfies the differential equations at a number of given points, namely, the collocation points [33, 34]. Owed to its efficiency and high accuracy, the collocation method has received many applications, including but not limited to fluid dynamics [35, 36], optimal control [37-39] and orbit propagation problems [40, 41]. The work [42] proposes a Chebyshev collocation-based batch state and dynamics estimation. However, it is more like trajectory optimization [39] than the general state estimation with priori information, covariance propagation and probabilistic interpretation [43]. Besides, the dynamics constraints at the discrete Chebyshev points therein are not optimal in the MAP sense and the batch method is not friendly to real-time applications.

The current paper is inspired by a series of our group works [44-47], in which the Chebyshev polynomial has been used to fully represent the rigid motion including attitude, velocity and position, and to numerically solve the involved motion dynamics, namely the forward propagation counterpart in state estimation. The motivation of this paper is to extend the Chebyshev's representation to incorporate the measurements and consequently to better solve the state estimation problem. Specifically, the continuous-time state in the time interval of interest is to be represented by a Chebyshev polynomial, and the continuous MAP state estimation is then transformed into a problem of Chebyshev coefficients optimization through the Chebyshev collocation method, named as the ChevOpt in this paper. Comparing with other kinds of polynomials, the Chebyshev polynomial is more accurate and very close to the best polynomial in the $\infty$-norm [34]. To meet the requirement of real-time applications, we further come up with a recursive sliding-window version (named as the W-ChevOpt), inspired by the strategy in MHE. It is noted that the model and noise are assumed well known and the influence of the model error and uncertainty [48, 49] is not considered in the current paper, although they should be seriously treated in any practical systems.

The main contribution of this work rests on a novel framework of continuous-time state estimation based on the Chebyshev polynomial in both batch and sliding-window forms. Comparing with well-known Gaussian filters, the proposed estimation does not introduce any approximation of the nonlinear dynamics and measurements and thus achieves remarkably improved accuracy. The remaining of the paper is organized as follows. Section II reviews the continuous-time batch MAP estimation problem. After a brief introduction of the Chebyshev collocation method, Section III proposes the optimal state estimation framework - ChevOpt. To meet the requirement of real-time applications, the W-ChevOpt is further proposed in Section IV that reformulates the batch ChevOpt in a sliding window with an additional covariance propagation algorithm. Section V is devoted to assessing the proposed algorithms and demonstrating their accuracy against traditional filter and smoother. The discussions and conclusions are finally drawn in Section VI.

## II. BATCH MAP ESTIMATION

The commonly-employed strategy of state estimation is maximum a posteriori, which is to find the most likely posterior state given the knowledge of prior, measurements and the dynamics model [13]. This section will review the probabilistic formulation of the batch MAP estimation for the nonlinear continuous-discrete system.

Consider a continuous-discrete nonlinear system, defined on the time interval $t \in [t_0 \ t_M]$ in the form of dynamic state-space model, as

$$\dot{\mathbf{x}}(t) = \mathbf{f}(\mathbf{x}(t), \mathbf{u}(t)) + \mathbf{G}(t)\mathbf{w}(t)$$
$$\mathbf{z}_k = \mathbf{h}(\mathbf{x}_k) + \mathbf{v}_k \tag{1}$$

where $\mathbf{x}(t) \in \mathbb{R}^n$ denotes the time-varying state, $\mathbf{z}_k \in \mathbb{R}^m$ denotes the measurement at time $t_k$. $\mathbf{f}(\bullet): \mathbb{R}^n \to \mathbb{R}^n$ and $\mathbf{h}(\bullet): \mathbb{R}^n \to \mathbb{R}^m$ are known continuous and smooth functions. $\mathbf{G}(t) \in \mathbb{R}^{n \times m}$ is the noise driven matrix, $\mathbf{u}(t) \in \mathbb{R}^p$ is the deterministic known control input, and $\mathbf{w}(t) \in \mathbb{R}^m$ is the zero-mean Gaussian dynamics noise with the covariance given by

$$E\{\mathbf{w}(t)\mathbf{w}^T(\tau)\} = \mathbf{Q}\delta(t-\tau) \tag{2}$$

where $\delta(\bullet)$ is the Dirac's delta function. $\mathbf{v}_k \sim N(\mathbf{0}, \mathbf{R}_k)$ is the measurement Gaussian noise, which is uncorrelated with $\mathbf{w}(t)$. Suppose that the priori state estimate $\mathbf{x}(t_0)$ is Gaussian-distributed, i.e., $\mathbf{x}(t_0) \sim N(\hat{\mathbf{x}}_0, \mathbf{P}_0)$. The goal of the MAP estimation is to find the best estimate for the state of the system $\mathbf{x}(t)$, given the prior information $\hat{\mathbf{x}}_0$, control input $\mathbf{u}$ and a sequence of measurements $\mathbf{z}_{1:M} \triangleq \{\mathbf{z}_1 \cdots \mathbf{z}_M\}$

$$\hat{\mathbf{x}}(t) = \arg\max_{\mathbf{x}} p(\mathbf{x} | \mathbf{u}, \mathbf{z}_{1:M}) \tag{3}$$

where $p(\mathbf{x} | \mathbf{u}, \mathbf{z}_{1:M})$ is the posterior probability density function (PDF) over the state. Using the Bayesian rule, (3) is equivalent to

$$\hat{\mathbf{x}}(t) = \arg\max_{\mathbf{x}} \frac{p(\mathbf{z}_{1:M} | \mathbf{x}, \mathbf{u}) p(\mathbf{x} | \mathbf{u})}{p(\mathbf{z}_{1:M} | \mathbf{u})}$$
$$\sim \arg\max_{\mathbf{x}} p(\mathbf{z}_{1:M} | \mathbf{x}, \mathbf{u}) p(\mathbf{x} | \mathbf{u}) \tag{4}$$

where $p(\mathbf{z}_{1:M} | \mathbf{x}, \mathbf{u})$ is the likelihood function of measurements and $p(\mathbf{x} | \mathbf{u})$ denotes the evolution of the state. The denominator in (4) is dropped because it does not depend on $\mathbf{x}$ [3, 13].

Define the weighted residuals of initial state, measurements and dynamics, respectively, as

$$\mathbf{e}_{\mathbf{x}_0} = \mathbf{W}_{\mathbf{x}_0}^T \left( \mathbf{x}(t_0) - \hat{\mathbf{x}}_0 \right)$$
$$\mathbf{e}_{\mathbf{z}_k} = \mathbf{W}_{\mathbf{z}_k}^T \left( \mathbf{z}_k - \mathbf{h}(\mathbf{x}_k) \right) \quad (5)$$
$$\mathbf{e}_{\mathbf{v}}(t) = \mathbf{W}_{\mathbf{v}}^T \left( \dot{\mathbf{x}}(t) - \mathbf{f}(\mathbf{x}(t), \mathbf{u}(t)) \right)$$

where $\mathbf{W}_{\mathbf{x}_0}$, $\mathbf{W}_{\mathbf{z}_k}$ and $\mathbf{W}_{\mathbf{v}}$ are lower triangular matrixes, obtained by the Cholesky factorization of $\mathbf{P}_0^{-1}$, $\mathbf{R}_k^{-1}$ and $(\mathbf{GQG})^{-1}$, that is to say, $\mathbf{P}_0^{-1} = \mathbf{W}_{\mathbf{x}_0} \mathbf{W}_{\mathbf{x}_0}^T$, $\mathbf{R}_k^{-1} = \mathbf{W}_{\mathbf{z}_k} \mathbf{W}_{\mathbf{z}_k}^T$ and $(\mathbf{GQG}^T)^{-1} = \mathbf{W}_{\mathbf{v}} \mathbf{W}_{\mathbf{v}}^T$.

Because the noises $\mathbf{v}_k$ and $\mathbf{w}(t)$ are uncorrelated, the likelihood function of the measurements can be simplified as

$$p(\mathbf{z}_{1:M} \mid \mathbf{x}, \mathbf{u}) = \eta_1 \prod_{k=1}^{M} \exp \left\{ -\frac{1}{2} \mathbf{e}_{\mathbf{z}_k}^T \mathbf{e}_{\mathbf{z}_k} \right\} \quad (6)$$

where $\eta_1$ is the normalization constant factor. And the probability density of $p(\mathbf{x} \mid \mathbf{u})$ in (4) is written as [43]

$$p(\mathbf{x} \mid \mathbf{u}) = \eta_2 p(\mathbf{x}_0) \exp \left\{ -\frac{1}{2} \int_{t_0}^{t_M} \mathbf{e}_{\mathbf{v}}(t)^T \mathbf{e}_{\mathbf{v}}(t) dt \right\} \quad (7)$$

where $\eta_2$ is the normalization constant factor. Substituting (6) and (7) into (4) and taking the negative logarithm of the probability density functions, the MAP estimation is transformed to the following optimization problem

$$\hat{\mathbf{x}}(t) = \arg\min_{\mathbf{x}} \left( J_{\mathbf{x}_0} + J_{\mathbf{z}_{1:M}} + J_{\mathbf{v},[t_0,t_M]} \right) \quad (8)$$

where $J_{\mathbf{x}_0}$, $J_{\mathbf{z}_{1:M}}$ and $J_{\mathbf{v},[t_0,t_M]}$ are defined as

$$J_{\mathbf{x}_0} = \mathbf{e}_{\mathbf{x}_0}^T \mathbf{e}_{\mathbf{x}_0}$$
$$J_{\mathbf{z}_{1:M}} = \sum_{k=1}^{M} \mathbf{e}_{\mathbf{z}_k}^T \mathbf{e}_{\mathbf{z}_k} \quad (9)$$
$$J_{\mathbf{v},[t_0,t_M]} = \int_{t_0}^{t_M} \mathbf{e}_{\mathbf{v}}(t)^T \mathbf{e}_{\mathbf{v}}(t) dt$$

The estimation problem in (8) is an infinite-dimension functional optimization, which can be transformed into a finite parameter optimization problem by the Chebyshev collocation method to be discussed in the sequel.

It is noted that in the computation of weighted matrices in (5), the dynamics-related term $\mathbf{GQG}^T$ is assumed to be positive definite, which may not be satisfied in many applications. This issue can be handled by dividing the dynamics into the sub-dynamics with a positive definite noise matrix and a noise-free sub-dynamics [47] (See details in the Appendix). Furthermore, there are several strategies to handle the issue to be discussed in detail in Section IV.

III. BATCH MAP ESTIMATION VIA CHEBYSHEV POLYNOMIAL

A. Chebyshev Polynomial and Collocation Method

The Chebyshev polynomial basis is defined over the interval [−1  1] by the recurrence relation as

$$F_0(\tau) = 1, \ F_1(\tau) = \tau,$$
$$F_{i+1}(\tau) = 2\tau F_i(\tau) - F_{i-1}(\tau) \ \text{for } i \geq 1 \quad (10)$$

where $F_i(\tau)$ is the $i^{th}$-degree Chebyshev polynomial of the first kind. The derivative of the Chebyshev polynomial is obtained by differentiating (10), as

$$\dot{F}_0(\tau) = 0, \ \dot{F}_1(\tau) = 1,$$
$$\dot{F}_{i+1}(\tau) = 2F_i(\tau) + 2\tau \dot{F}_i(\tau) - \dot{F}_{i-1}(\tau) \ \text{for } i \geq 1 \quad (11)$$

where $\dot{F}_i(\tau)$ denotes the derivative of the $i^{th}$-degree Chebyshev polynomial. And the integrated $i^{th}$-degree Chebyshev polynomial can be expressed as a linear combination of $(i+1)^{th}$-degree Chebyshev polynomial, given by [50]

$$G_{i,[-1,\tau]} = \int_{-1}^{\tau} F_i(\tau) d\tau = \begin{cases} \left( \frac{F_{i+1}(\tau)}{2(i+1)} - \frac{F_{i-1}(\tau)}{2(i-1)} \right) - \frac{(-1)^i}{i^2-1} F_0(\tau), & i \neq 1 \\ \frac{F_{i+1}(\tau)}{4} - \frac{F_0(\tau)}{4}, & i = 1 \end{cases} \quad (12)$$

Consider a nonlinear differential equation on $t \in [a \ b]$

$$f(\dot{y}(t), y(t), t) = 0 \quad (13)$$

where $f(\cdot)$ is a smooth function and $y$ is the solution of interest. In order to use the Chebyshev polynomial, the original interval is mapped into $\tau \in [-1 \ 1]$ by the affine transformation

$$\tau = \frac{2}{b-a} t - \frac{b+a}{b-a} \quad (14)$$

Then, the function $y$ can be approximated by the Chebyshev polynomial of degree $N$, together with its derivative $\dot{y}$, as

$$y(\tau) \approx y_N(\tau) = \sum_{i=0}^{N} c_i F_i(\tau) \quad (15)$$

$$\dot{y}(\tau) \approx \dot{y}_N(\tau) = \sum_{i=0}^{N} c_i \dot{F}_i(\tau) \quad (16)$$

where $c_i$ is the Chebyshev coefficient to be determined. The collocation method [33] approximates the solution $y(\tau)$ by the polynomial $y_N(\tau)$ of degree $N$, which satisfies the differential equaion at the chosen collocation points $\tau_k \in [-1 \ 1]$ for $k = 0, \cdots, N$, as follows

$$f(\dot{y}(\tau_k), y(\tau_k), \tau_k) = 0 \quad k = 0, 1, \cdots, N \quad (17)$$

Generally, the collocation points associated with Chebyshev polynomial are taken as the Chebyshev points, i.e.,

$$\tau_k = -\cos(k\pi/N), \quad k = 0, 1, \cdots, N \quad (18)$$

Substituting (15), (16) and (18) into (17), the nonlinear differential equation is transformed into nonlinear equations about the Chebyshev coefficient $c_k$, which can be solved by many optimization algorithms, e.g., the Newton method [33]. Interested readers are referred to [33, 51] and references therein for more details about the collocation method.

Instead of estimating the Chebyshev coefficients in (16), the objective function can also be constructed using the unknown states at the collocation points [52]. In fact, the function $u$ can be equally represented by its value at each collocation point or the Chebyshev coefficients, and the transformation of these two representations can be realized efficiently by the Fast Fourier Transform [34].

*B. Batch MAP Estimation Via Collocation Method*

The objective function for the batch MAP estimation in (8) can be solved by the Chebyshev collocation method. Transforming the time interval $[t_0 \;\; t_M]$ to $[-1 \;\; 1]$ by (14) and approximating the state and its derivative by Chebyshev polynomials as

$$\mathbf{x}(\tau) \approx \mathbf{x}_N(\tau) \triangleq \sum_{i=0}^{N} \mathbf{d}_i F_i(\tau) \quad (19)$$

$$\dot{\mathbf{x}}(\tau) \approx \dot{\mathbf{x}}_N(\tau) = \sum_{i=0}^{N} \mathbf{d}_i \dot{F}_i(\tau) \quad (20)$$

where $\mathbf{d}_i \in \mathbb{R}^n$ is the $i^{\text{th}}$-degree Chebyshev coefficient to be determined. With these approximations, the batch MAP estimation in (8) is transformed into the parameter optimization problem as

$$\mathbf{d}_{0:N} = \arg\min_{\mathbf{d}_{0:N}} \left( J'_{\mathbf{x}_0} + J'_{\mathbf{z}_{1:M}} + J'_{\mathbf{v},[-1,1]} \right) \quad (21)$$

where $\mathbf{d}_{0:N} \triangleq \left[\mathbf{d}_0^T, \cdots, \mathbf{d}_N^T\right]^T$ and the terms $J'_{\mathbf{x}_0}$, $J'_{\mathbf{z}_{1:M}}$ and $J'_{\mathbf{v},[-1,1]}$ are defined as

$$J'_{\mathbf{x}_0} = \mathbf{e}'^T_{\mathbf{x}_0} \mathbf{e}'_{\mathbf{x}_0}$$

$$J'_{\mathbf{z}_{1:N}} = \sum_{k=1}^{M} \mathbf{e}'^T_{\mathbf{z}_k} \mathbf{e}'_{\mathbf{z}_k} \quad (22)$$

$$J'_{\mathbf{v},[-1,1]} = \frac{t_M - t_0}{2} \int_{-1}^{1} \mathbf{e}'_{\mathbf{v}}(\tau)^T \mathbf{e}'_{\mathbf{v}}(\tau) d\tau$$

where $\mathbf{e}'_{\mathbf{x}_0}$, $\mathbf{e}'_{\mathbf{z}_k}$ and $\mathbf{e}'_{\mathbf{v}}$ are the weighted residuals of initial state, measurements and dynamics, respectively, in the form of Chebyshev polynomials

$$\mathbf{e}'_{\mathbf{x}_0} = \mathbf{W}^T_{\mathbf{x}_0} \left( \mathbf{x}_N(-1) - \hat{\mathbf{x}}_0 \right)$$

$$\mathbf{e}'_{\mathbf{z}_k} = \mathbf{W}^T_{\mathbf{z}_k} \left( \mathbf{z}_k - \mathbf{h}(\mathbf{x}_N(\tau_k)) \right) \quad (23)$$

$$\mathbf{e}'_{\mathbf{v}}(\tau) = \mathbf{W}^T_{\mathbf{v}} \left( \frac{2}{t_M - t_0} \dot{\mathbf{x}}_N(\tau) - \mathbf{f}(\mathbf{x}_N(\tau), \mathbf{u}(\tau)) \right)$$

The integrated term $J_{\mathbf{v},[-1,1]}$ in (21) can be approximated by the Clenshaw-Curtis quadrature [34] formula as

$$J'_{\mathbf{v},[-1,1]} \approx \frac{t_M - t_0}{2} \sum_{i=0}^{N} \omega_i \mathbf{e}'_{\mathbf{v}}(\tau_i)^T \mathbf{e}'_{\mathbf{v}}(\tau_i) \quad (24)$$

where $\omega_i$ is the weight to be predetermined by the integral of the Lagrange polynomial

$$\omega_i = \int_{-1}^{1} \ell_i(\tau) d\tau \quad (25)$$

and the Lagrange polynomial is given as

$$\ell_i(\tau) = \prod_{k=0, k\neq i}^{N} (\tau - \tau_i) \Big/ \prod_{k=0, k\neq i}^{N} (\tau_k - \tau_i) \quad (26)$$

Substituting (24) into (21), the optimization problem is transformed into a nonlinear least squares problem about $\mathbf{d}_{0:N}$, which can be solved by a number of algorithms, e.g. the Levenberg-Marquardt method [53]. With the optimized parameter $\mathbf{d}_{0:N}$, the state over the whole time interval is then acquired by (19).

The state approximation by the Chebyshev polynomial in (19) and the Clenshaw-Curtis quadrature in (24) are the only two approximations in the proposed ChevOpt framework. As for state approximation, the Chebyshev polynomial is very close to the best polynomial approximation in the $\infty$-norm [34]. On the other hand, the Clenshaw-Curtis quadrature converges for any continuous function [54]. With the increased order of Chebyshev polynomial $N$, these approximation errors will approach zero. It means that the proposed ChevOpt is optimal in the least squares sense insofar as the number of collocation points are sufficiently large.

IV. W-CHEVOPT: RECURSIVE MAP ESTIMATION VIA CHEBYSHEV POLYNOMIAL

Since the batch ChevOpt can only be implemented until all the measurements come in, we turn to solve the MAP estimation by the ChevOpt in a sliding time window.

Assume the current sliding window is $[t_{m-l} \;\; t_m]$, where $l$ is the size of the sliding window. Following (21)-(23), the MAP estimation in the sliding window can be readily formulated as

$$\mathbf{d}_{0:N} = \arg\min_{\mathbf{d}_{0:N}} \left( J'_{\mathbf{x}_{m-l}} + J'_{\mathbf{z}_{m-l+1:m}} + J'_{\mathbf{v},[-1,1]} \right) \quad (27)$$

where $J'_{\mathbf{x}_{m-l}}$ is the state prior including and before $t_{m-l}$, and the terms are defined as follows

$$J'_{\mathbf{x}_{m-l}} = \mathbf{e}'^T_{\mathbf{x}_{m-l}} \mathbf{e}'_{\mathbf{x}_{m-l}}$$

$$J'_{\mathbf{z}_{m-l+1:m}} = \sum_{k=m-l+1}^{m} \mathbf{e}'^T_{\mathbf{z}_k} \mathbf{e}'_{\mathbf{z}_k} \quad (28)$$

$$J'_{\mathbf{v},[-1,1]} = \frac{t_m - t_{m-l}}{2} \int_{-1}^{1} \mathbf{e}'_{\mathbf{v}}(\tau)^T \mathbf{e}'_{\mathbf{v}}(\tau) d\tau$$

where

$$\mathbf{e}'_{\mathbf{x}_{m-l}} = \mathbf{W}^T_{\mathbf{x}_{m-l}} \left( \mathbf{x}_N(-1) - \hat{\mathbf{x}}_{m-l} \right)$$

$$\mathbf{e}'_{\mathbf{z}_k} = \mathbf{W}^T_{\mathbf{z}_k} \left( \mathbf{z}_k - \mathbf{h}(\mathbf{x}_N(\tau_k)) \right) \quad (29)$$

$$\mathbf{e}'_{\mathbf{v}}(\tau) = \mathbf{W}^T_{\mathbf{v}} \left( \frac{2}{t_m - t_{m-l}} \dot{\mathbf{x}}_N(\tau) - \mathbf{f}(\mathbf{x}_N(\tau), \mathbf{u}(\tau)) \right)$$

The weight matrix $\mathbf{W}_{\mathbf{x}_{m-l}}$ is obtained by the Cholesky factorization of $\mathbf{P}^{-1}_{m-l}$ as $\mathbf{P}^{-1}_{m-l} = \mathbf{W}_{\mathbf{x}_{m-l}} \mathbf{W}^T_{\mathbf{x}_{m-l}}$.

To acquire the initial covariance $\mathbf{P}_{m-l}$ at the beginning of a sliding window, a convenient strategy is to approximate the prior as a Gaussian distribution and use the extended Kalman filter for covariance prediction and update [26] with the state estimated by the W-ChevOpt. Specifically, for a continuous-discrete system, the EKF prediction covariance is governed by [13]

$$\dot{\mathbf{P}}(t) = \mathbf{F}(t)\mathbf{P}(t) + \mathbf{P}(t)\mathbf{F}^T(t) + \mathbf{Q}(t) \quad (30)$$

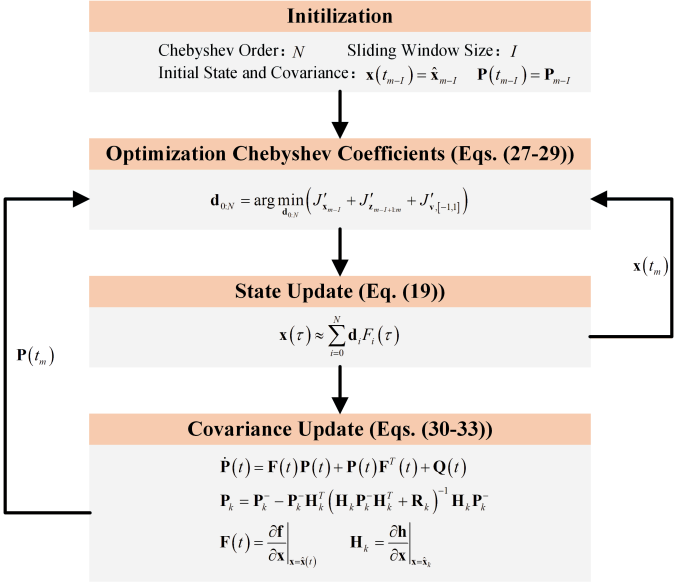

Fig. 1. Flowchart of the W-ChevOpt.

where $\mathbf{F}(\cdot)$ is the Jacobian function of dynamics. Instead of using the prediction state by the EKF, $\mathbf{F}(\cdot)$ is evaluated at the estimated state by the W-ChevOpt

$$\mathbf{F}(t) = \left.\frac{\partial \mathbf{f}}{\partial \mathbf{x}}\right|_{\mathbf{x}=\hat{\mathbf{x}}(t)} \tag{31}$$

The integral of (30) can be solved by either the traditional Runge-Kutta or the Chebyshev collocation method as well. When it comes to the measurement at time $t_k$, the predicted covariance is updated by

$$\mathbf{P}_k = \mathbf{P}_k^- - \mathbf{P}_k^- \mathbf{H}_k^T \left(\mathbf{H}_k \mathbf{P}_k^- \mathbf{H}_k^T + \mathbf{R}_k\right)^{-1} \mathbf{H}_k \mathbf{P}_k^- \tag{32}$$

where $\mathbf{P}_k^-$ is the predicted covariance at time $t_k$ by (30), and $\mathbf{H}_k$ is the Jacobian of $\mathbf{h}(\cdot)$ also evaluated at the estimated state in the sliding window by the W-ChevOpt

$$\mathbf{H}_k = \left.\frac{\partial \mathbf{h}}{\partial \mathbf{x}}\right|_{\mathbf{x}=\hat{\mathbf{x}}_k} \tag{33}$$

It is noted that the W-ChevOpt state estimation in (27) utilizes all the measurements in the window while the covariance computation based on the EKF (with Jacobians evaluated at the ChevOpt estimated state) uses the measurements only up to the current time. To improve the covariance consistency with the state, a covariance smoother, e.g. extended RTS smoother [13, 14] with Jacobian evaluated at the ChevOpt estimated state in the sliding window, can be used. However, for those applications that only consider the state estimation, the covariance smoother may be unnecessary, as the ChevOpt only requires the covariance at the end of current window as the prior covariance for the next window. It is well known that the result of a smoother at the end of current window is the same with that of a filter.

Figure 1 lists the main steps of the W-ChevOpt. The estimation is initialized with the user-given Chebyshev order, sliding window size, initial state and corresponding covariance. The Chebyshev coefficients of the state are estimated by minimizing the objective function in (27) and the initial coefficients of the optimization can be acquired by the W-ChevOpt with a smaller window size or the fitting polynomial of the EKF result. After the optimization, the continuous state over the sliding window is obtained by the Chebyshev polynomial approximation. And, the corresponding covariance is calculated by the EKF with the linearized dynamics and measurement at the W-ChevOpt estimated state. Finally, the estimated state and covariance at the end of the sliding window are treated as the prior for the next sliding window.

In the case of semi-positive definite $\mathbf{GQG}^T$ as discussed in Section II and Appendix, the dynamics can be divided into the sub-dynamics with a positive definite noise matrix and the noise-free sub-dynamics. The proposed ChevOpt framework can be used on the positive definite sub-dynamics directly. And, for the noise-free sub-dynamics, there are several strategies to handle it as follows:

- If the noise-free sub-dynamics function $\mathbf{g}^{(2)}(\mathbf{y}(t), \mathbf{u}(t))$ in (50) does not depend on the noise-free state $\mathbf{y}^{(2)}$ at all, i.e., $\mathbf{g}^{(2)}(\mathbf{y}(t), \mathbf{u}(t)) \equiv \mathbf{g}^{(2)}(\mathbf{y}^{(1)}(t), \mathbf{u}(t))$, $\mathbf{y}^{(2)}$ can be removed from the estimation and represented as a function of the remaining state $\mathbf{y}^{(1)}$. (See Examples 1-2 in Section V)
- Remove the noise-free part of the dynamics from the objective function and treat it instead as a constraint in the optimization (21) or (27).
- Add a small disturbance to the noise-free sub-dynamics to make the noise covariance positive definite (See Example 2 about the velocity $x_2$). This strategy has been commonly used in the traditional state estimation methods, see e.g., [16] and [55].

Note that the first strategy helps decrease the state space dimensionality, which is preferable than the other two strategies in reducing computation burden. The third strategy is approximate in terms of adding pseudo-noises, while the second strategy is most time-consuming due to the constrained optimization.

## V. SIMULATIONS AND RESULTS

In this section, we will evaluate the proposed batch ChevOpt and recursive W-ChevOpt by two representative estimation examples, viz. the Van der Pol model and the ballistic target reentry, which have been extensively investigated in the literature [9, 14, 15, 56], and compare with the continuous-time EKF, UKF, batch extended Rauch, Tung, and Striebel (ERTS) smoother and fixed-lag ERTS. It is noted that although both the W-ChevOpt and fixed-lag ERTS are sliding window-based estimators, the window sliding step of the fixed-lag ERTS equals the measurement sampling interval while that of the W-ChevOpt is the sliding window size and its adjacent windows do not overlap. At the end time of each sliding window, the W-ChevOpt utilizes the measurements only up to the current time, while the fixed-lag ERTS needs to wait for future measurements to come in for estimation.

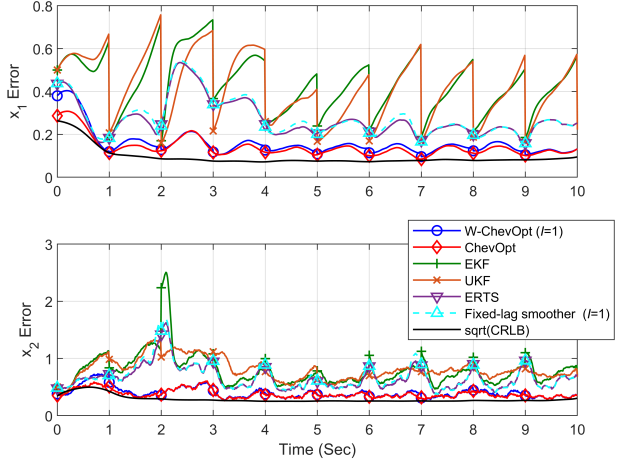

Fig. 2. Averaged absolute errors of the state (the W-ChevOpt and fixed-lag ERTS with window size of 1s) and the square root of CRLB.

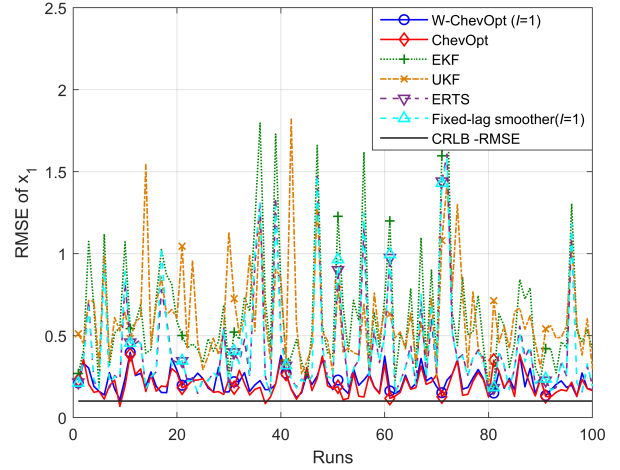

Fig. 3. RMSE of $x_1$ (the W-ChevOpt and fixed-lag ERTS with window size of 1s) across 100 random runs and CRLB-RMSE.

All the mentioned algorithms in this paper are performed on the MATLAB platform and the function 'lsqnonlin' with analytical Jocabian is used to solve the nonlinear optimization in both the ChevOpt and W-ChevOpt. Besides, the Cramer-Rao lower bound (CRLB) [57, 58] for batch smoother is presented as the reference of performance comparison, as it sets a theoretical limit on the best achievable performance of any estimator, which admits us to quantify how much scope is left to improve estimation [8].

A. *Example 1: Van der Pol Oscillator*

The Van der Pol's equation is defined in the state-space model as [56]
$$\dot{x}_1 = x_2 + w_1 \\ \dot{x}_2 = -\lambda(x_1^2 - 1)x_2 - x_1 + w_2 \tag{34}$$

where $[w_1 \ w_2]^T \sim N(\mathbf{0}, \mathbf{Q}_c)$ is the continuous dynamics noise and $\mathbf{x} \triangleq [x_1 \ x_2]^T$ is the state vector. The parameter $\lambda$ is a constant value ($\lambda = 3$). The measurement equation at time $t_k$ is given as
$$z_k = x_1(t_k) + v_k \tag{35}$$
where $v_k \sim N(0, R_k)$ is the white measurement noise and the frequency of the measurement is assumed to be 1 Hz.

The spectral density of the continuous dynamics noise is $\mathbf{Q}_c = \begin{bmatrix} 0 & 0 \\ 0 & 1 \end{bmatrix}$ and the measurement noise covariance is $R_k = 0.04$. The reference in the simulation is generated over a time period of 10 seconds using the Euler-Maruyama method [59] at the time step $\Delta t = 5 \times 10^{-4}$ s, with the true initial state $\mathbf{x}_0 = [0.5 \ 0.5]^T$. The relationship between the spectral density and the discrete covariance is approximated as $\mathbf{Q}_d \approx \mathbf{Q}_c \Delta t$.

For all estimators, the initial state is $\hat{\mathbf{x}}_0 = [1 \ 1]^T$ with the covariance
$$P_0 = \begin{bmatrix} 0.25 & 0 \\ 0 & 0.25 \end{bmatrix}$$
The step size of prediction for the EKF, UKF and ERTS smoother is 0.01s.

In this example, the first strategy listed in Section IV is applied to handle the semi-positive dynamics noise covariance, where the state component $x_2$ is approximated by the Chebyshev polynomial and $x_1$ is obtained by the integral of $x_2$
$$x_1(\tau) \approx p_0 + \sum_{i=0}^{N} c_i G_{i,[-1,\tau]} \\ x_2(\tau) \approx \sum_{i=0}^{N} c_i F_i(\tau) \tag{36}$$
where $c_i$ is the Chebyshev coefficient to be determined, $p_0$ is the initial value of $x_1$ at the left end of sliding window, and $G_{i,[-1,\tau]}$ is the integral of Chebyshev polynomial $F_i(\tau)$ as given in (12). By so doing, the semi-positive definite dynamics noise covariance is naturally circumvented and the state dimensionality of the nonlinear optimization is also reduced by a half.

The estimation parameters in this example $\mathbf{d} \triangleq \begin{bmatrix} p_0 & c_{0:N}^T \end{bmatrix}$ include the Chebyshev coefficients and initial position, which are determined by substituting (36) into (21) for the ChevOpt or (27) for the W-ChevOpt, respectively. The time window sizes of the W-ChevOpt and fixed-lag ERTS are both set to 1 second. The Chebyshev polynomial orders $N$ are set to 300 and 20 for the ChevOpt and W-ChevOpt, respectively.

As per (34), the CRLB computation requires an invertible dynamics noise covariance. A small value is assumed for $w_1$, namely, $w_1 \sim N(0, 0.001^2)$, to make the dynamics noise

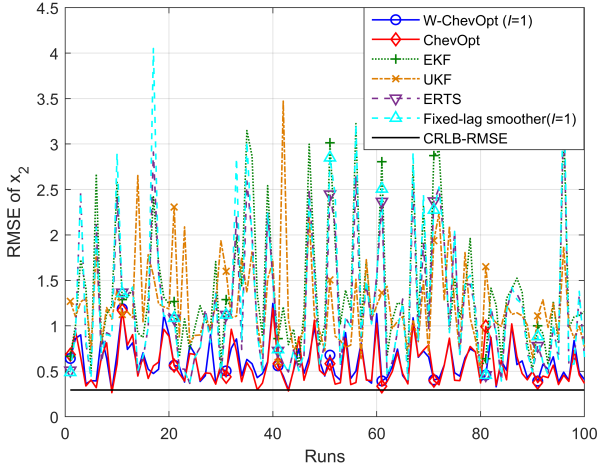

Fig. 4. RMSE of $x_2$ (the W-ChevOpt and fixed-lag ERTS with window size of 1s) across 100 random runs and CRLB-RMSE.

TABLE I. ACCUMULATIVE RMSE OF EXAMPLE #1

|  | $x_1$ | $x_2$ |
| --- | --- | --- |
| W-ChevOpt ($l$=1) | 0.23 | 0.64 |
| ChevOpt | 0.22 | 0.62 |
| EKF | 0.73 | 1.55 |
| UKF | 0.68 | 1.31 |
| ERTS | 0.52 | 1.37 |
| Fixed-Lag ERTS ($l$=1) | 0.54 | 1.48 |
| CRLB-RMSE | 0.10 | 0.29 |

covariance invertible. The averaged absolute error of $i$-th state component at time $k$ is defined by

$$\varepsilon_i(k) = \frac{1}{L}\sum_{l=1}^{L}\left|\hat{\mathbf{x}}_{(i),k}^{(l)} - \mathbf{x}_{(i),k}^{(l)}\right| \quad (37)$$

where $L$ is the number of Monte Carlo runs. Note that the performance evaluation indicators in the sequel include the averaged absolute error and the root mean square error, following the works [8, 16, 18]. Figure 2 plots the averaged absolute state errors of 100 Monte Carlo runs by different estimation algorithms and the square root of CRLB. We see that the W-ChevOpt and ChevOpt have the best accuracy, with the former being slightly worse than the latter. The list of the estimation accuracy is

$$\text{CRLB} > \text{ChevOpt} > \text{W-ChevOpt} > \begin{cases} \text{UKF} \\ \text{ERTS} > \text{Fixed-lag ERTS} \end{cases} > \text{EKF} \quad (38)$$

As we indicated in the introduction that the EKF/ERTS smoother and UKF make approximations about the dynamics and measurement nonlinearities by the first-order Taylor expansion and the deterministic sampling, respectively. In contrast, the proposed ChevOpt is an optimal state estimator in the MAP sense and the W-ChevOpt only makes a Gaussian assumption of the priori distribution. As a result,

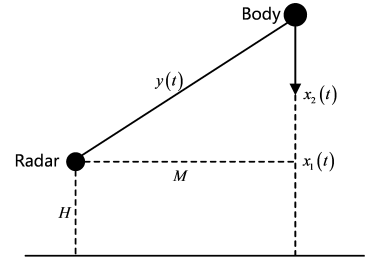

Fig. 5. Geometry of the ballistic target re-entry [14]

TABLE II. COMPARISON OF COMPUTATION COST OF EXAMPLE #1

|  | W-ChevOpt | ChevOpt | EKF | UKF | ERTS | Fixed-Lag ERTS |
| --- | --- | --- | --- | --- | --- | --- |
| Time Cost (s) | 0.07 | 0.81 | 0.02 | 0.10 | 0.16 | 0.17 |

they achieve remarkably improved accuracy over the EKF/ERTS smoother and UKF.

The performance of different algorithms is also compared in terms of the root mean square error (RMSE), which is defined for $l$-th Monte Carlo run by

$$RMSE(l) = \sqrt{\frac{1}{K}\sum_{k=1}^{K}\left(\hat{\mathbf{x}}_k^{(l)} - \mathbf{x}_k^{(l)}\right)^2} \quad (39)$$

where $K$ is the number of time instants. And, the CRLB RMSE is defined by the root of averaging CRLB across $K$ time instants, i.e.,

$$CRLB\text{-}RMSE = \sqrt{\frac{1}{K}\sum_{k=1}^{K}CRLB(k)} \quad (40)$$

Figures 3-4 plot the RMSE of the estimated state and CRLB across 100 Monte Carlo runs. The accumulative RMSE for each estimator is listed in Table I, which is calculated by

$$ARMSE = \sqrt{\frac{1}{LK}\sum_{l=1}^{L}\sum_{k=1}^{K}\left(\hat{\mathbf{x}}_k^{(l)} - \mathbf{x}_k^{(l)}\right)^2} \quad (41)$$

These results clearly show the accuracy superiority of the ChevOpt and W-ChevOpt over the EKF/UKF and batch/fixed-lag ERTS smoother.

The average time cost of 100 Monte Carlo runs is listed in Table II. The time cost of the W-ChevOpt is about 3 times larger than that of the EKF, yet still more efficient than UKF. Although the batch ChevOpt is about 40 times slower than the EKF, it is intended for post-processing in lieu of real-time applications.

*B. Example 2: Ballistic Target Re-entry*

This example is to estimate the altitude, velocity and ballistic coefficient of a vertically falling body as it re-enters the atmosphere at a very high altitude with high speed [60]. The geometry of this problem is shown in Fig. 5, where $x_1(t)$ denotes the altitude, $x_2(t)$ denotes the falling velocity, and $y(t)$ is the measured distance between the radar and the body. The altitude of the radar is $H = 10^5$ feet and the horizontal distance between radar and the body is $M = 10^5$ feet.

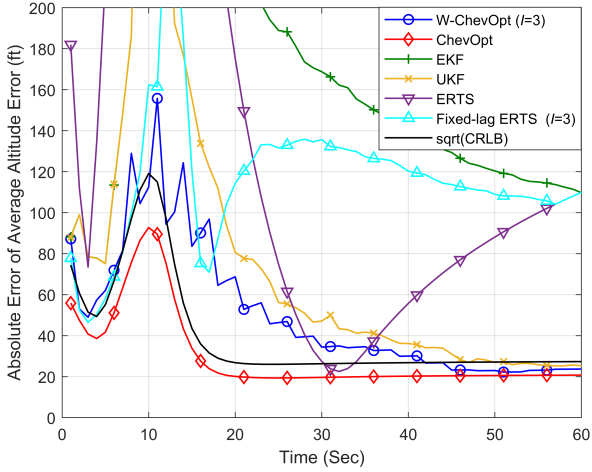

Fig. 6. Averaged absolute position errors (the W-ChevOpt and fixed-lag ERTS with window size 3s) and the square root of CRLB.

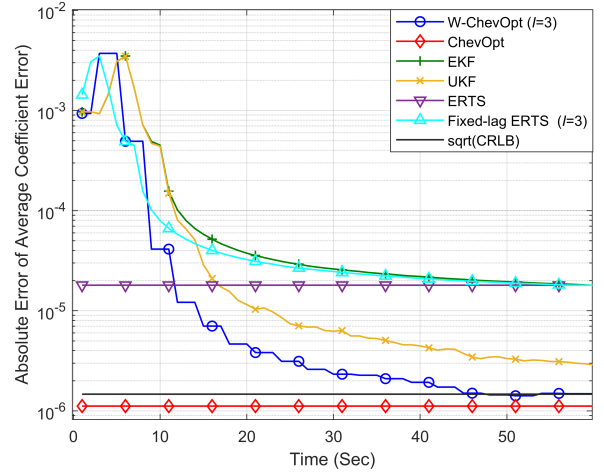

Fig. 8. Averaged absolute ballistic coefficient errors (the W-ChevOpt and fixed-lag ERTS with window size 3s) and the square root of CRLB.

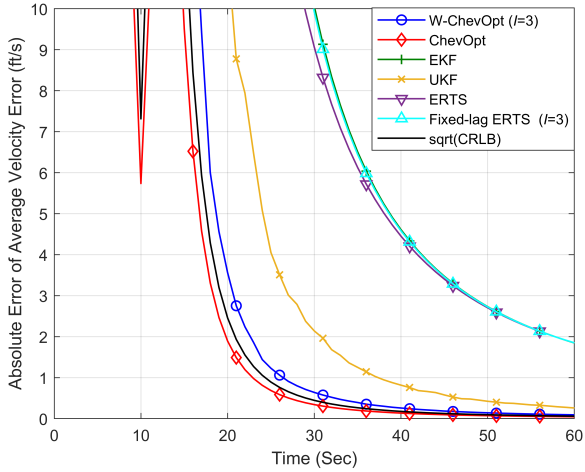

Fig. 7. Averaged absolute velocity errors (the W-ChevOpt and fixed-lag ERTS with window size 3s) and the square root of CRLB.

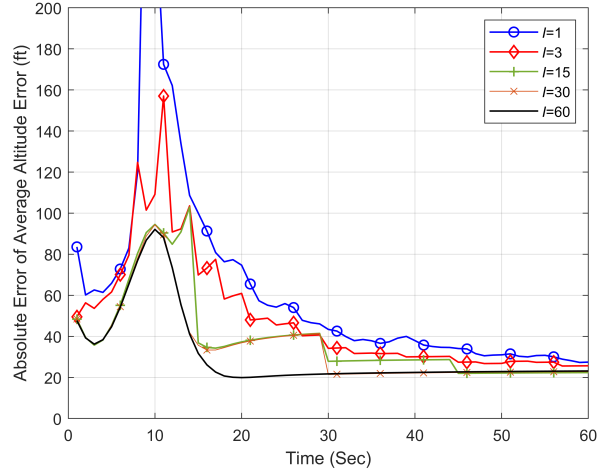

Fig. 9. Averaged absolute position errors of the W-ChevOpt with different window sizes (1s, 3s, 15s, 30s and 60s).

The continuous dynamic model is given as

$$\begin{aligned}\dot{x}_1(t) &= -x_2(t) + w_1(t) \\ \dot{x}_2(t) &= -e^{-\gamma x_1(t)} x_2^2(t) x_3(t) + w_2(t) \\ \dot{x}_3(t) &= w_3(t)\end{aligned} \quad (42)$$

where $[w_1(t)\ w_2(t)\ w_3(t)]^T \sim N(\mathbf{0}, \mathbf{Q})$ and $\gamma$ is a constant value ($\gamma = 5 \times 10^{-5}$) that relates the air density with altitude. The radar measurement at time $t_k$ is given as

$$y_k = \sqrt{(x_1(t_k) - H) + M^2} + v_k \quad (43)$$

where $v_k \sim N(0, R_k)$ and the measurement frequency is 1 Hz.

Following the previous literature [8, 14], the dynamic model has no noise, i.e., $\mathbf{Q} = 0$, and the covariance of measurement noise is set to $R_k = 10^4$. The true initial state is $\mathbf{x}_0 = [3 \times 10^5,\ 2 \times 10^4,\ 10^{-3}]^T$ and the initial state in the estimation is $\hat{\mathbf{x}}_0 = [3 \times 10^5,\ 2 \times 10^4,\ 3 \times 10^{-5}]^T$ with the covariance given as

$$P_0 = \begin{bmatrix} 10^6 & 0 & 0 \\ 0 & 4 \times 10^6 & 0 \\ 0 & 0 & 10^{-4} \end{bmatrix}$$

To address the issue of semi-positive definiteness, the altitude $x_1$ is represented using the velocity $x_2$'s Chebyshev coefficients as in Example 1 and the ballistic coefficient $x_3$ is taken as a constant. As for the second dynamics equation in (42), we add a small disturbance $w_2 \sim N(0, 0.001^2)$ to make it positive definite. Therefore, the estimation parameters of the W-ChevOpt and ChevOpt in this example are defined as $\mathbf{d} \triangleq [x_3\ p_0\ c_{0:N}^T]^T$, where $p_0$ and $c_{0:N}^T$ are defined in (36).

To overcome the high nonlinearity of the dynamic model in (42), the differential equations are integrated by a fourth-order Runge-Kutta scheme with a step size of $1/64$ seconds for both the EKF and UKF.

The averaged absolute errors of position, velocity and the ballistic coefficient by different estimators across 100 Monte Carlo runs are plotted in Figs. 6-8. The sliding window sizes of the W-ChevOpt and fixed-lag ERTS are 3 seconds. The Chebyshev polynomial orders of the W-ChevOpt and ChevOpt are set to 20 and 150, respectively. The list of state estimation accuracy is

$$\text{ChevOpt} > \text{CRLB} > \text{W-ChevOpt} > \begin{cases} \text{UKF} \\ \text{ERTS} > \text{Fixed-lag ERTS} \end{cases} > \text{EKF} \quad (44)$$

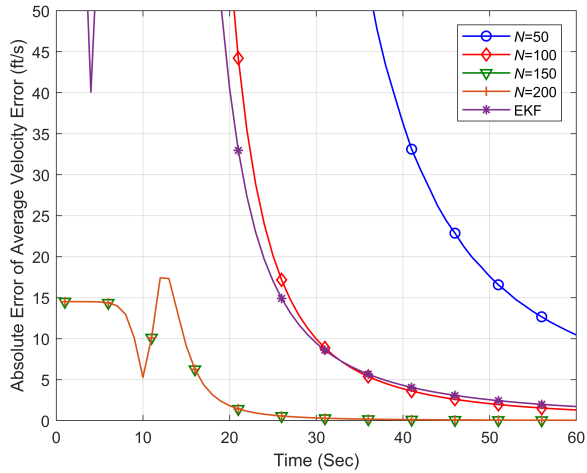

Fig. 10. Averaged absolute velocity errors of the ChevOpt with different Chebyshev orders ($N$=50, 100, 150, 200) and EKF.

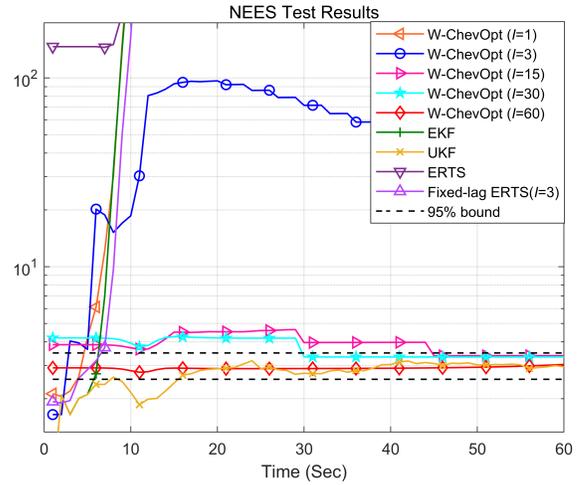

Fig. 12. Normalized estimation error squared (NEES) test of different estimators and the 95% consistency bound.

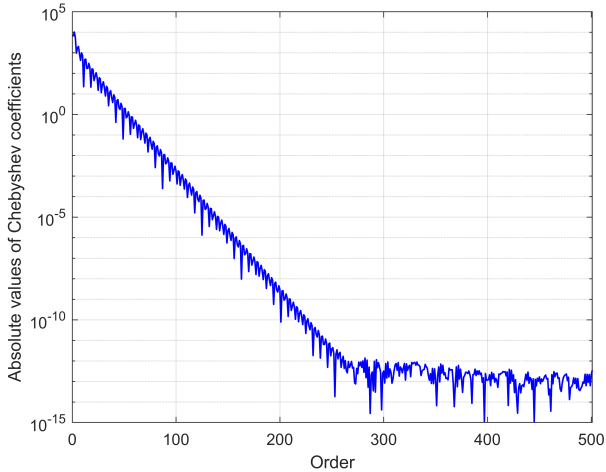

Fig. 11. Absolute values of Chebyshev coefficients of the ChevOpt with Chebyshev polynomial order 500.

It is surprising that the ChevOpt achieves even better accuracy than the CRLB does in this example. This is arguably because that the smoother CRLB for this noise-free dynamic model is evaluated by the covariance propagation of the ERTS with the Jacobians computed at the true state, and additionally the initial covariance is too large and inconsistent with the initial state in this example. Note that the initial altitude and velocity are true while the corresponding standard deviations are set to 1000 ft and 2000 ft/s following [8, 14]. Another reason may be that the nonlinearities of dynamics and measurement are better handled by the Chebyshev polynomial-based methods, as compared with the linearization model in the CRLB computation.

To evaluate the influence of the sliding window size, the averaged absolute altitude errors by the W-ChevOpt with different window sizes ($I$ = 1, 3, 15, 30 and 60 seconds) are shown in Figs. 9. The corresponding Chebyshev polynomial orders are set to 10, 20, 80, 100 and 150, respectively. It is shown that the estimation accuracy improves along with the increasing size of the sliding window. Note that the W-ChevOpt with the window size of 60 seconds is identical to the ChevOpt. With the increased window size, the effect of prior approximation in the optimization decreases and the W-ChevOpt approaches the ChevOpt. However, a larger window size brings about a longer time delay in carrying out the state estimation, as the estimation cannot be done until all the measurements in the sliding window come in. Therefore, the selection of the window size for the W-ChevOpt in real applications needs to consider the balance between the accuracy and the real-time requirement.

Another adjustable parameter that influences the accuracy of the estimation is the Chebyshev polynomial order to represent the state. A principle to set the Chebyshev polynomial order is to ensure that the Chebyshev polynomial approximation error is small enough in the state estimation. In the current implementation, the order of the Chebyshev polynomial requires to be manually prescribed, according to the dynamics of the model and the sliding window size. In general, high dynamics and large sliding window require a high-order Chebyshev polynomial to reduce the approximation error. But when the polynomial order is big enough, further increasing the order leads to heavy computer burden with little accuracy improvement. Figure 10 plots the averaged absolute velocity errors for the batch ChevOpt of different polynomial orders, in which the best performance is achieved when the Chebyshev polynomial order exceeds 150. A useful strategy to identify an appropriate polynomial order is by the convergence tendency and the absolute value of the estimated Chebyshev coefficients. Figure 11 plots the estimated Chebyshev coefficients of the ChevOpt with polynomial order 500. We see that the

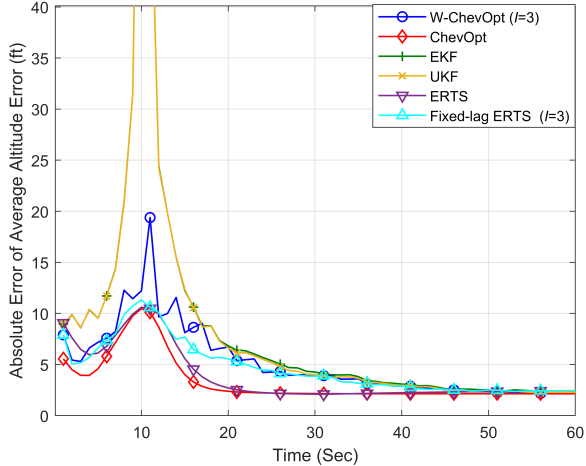

Fig. 13. Averaged absolute position errors with reduced measurement covariance (the W-ChevOpt and fixed-lag ERTS with window size 3s).

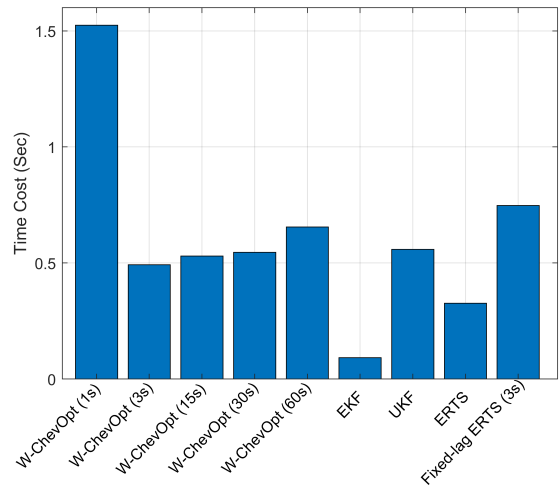

Fig. 14. Time costs of the W-ChevOpt with different window sizes (1s, 3s, 15s, 30s and 60s), EKF, UKF, ERTS and fixed-lag ERTS (3s).

Chebyshev coefficients converge exponentially before they reach the truncation error level at the order of about 270.

The consistency of different estimators is tested by the normalized estimation error squared (NEES) [14]

$$\zeta_k = \frac{1}{L}\sum_{l=1}^{L}\left(\hat{\mathbf{x}}_k^{(l)} - \mathbf{x}_k\right)^T \left(\mathbf{P}_k^{(l)}\right)^{-1} \left(\hat{\mathbf{x}}_k^{(l)} - \mathbf{x}_k\right) \quad (45)$$

where $\hat{\mathbf{x}}_k^{(l)}$ and $\mathbf{P}_k^{(l)}$ denote the estimated state and covariance at time $t_k$ for the $l$-th run, respectively. $\zeta_k$ satisfies a chi-square density with degree $Ln$ if the estimation is consistent ($L$ is the number of Monte Carlo runs and $n$ is the state dimension). Therefore, the consistency can be checked by the chi-square test. In this example, the consistency hypothesis is accepted with 95% confidence if $\zeta_k \in [2.53 \quad 3.49]$ for 100 Monte Carlo runs as shown in Fig. 12. It can be seen that the batch ChevOpt is consistent at all times, the UKF and W-ChevOpt with window sizes of 15s and 30s are also consistent after convergence, and the EKF, batch/fixed-lag ERTS and W-ChevOpt with window sizes of 1s and 3s are inconsistent in the whole-time interval. The inconsistency of the W-ChevOpt with small window sizes is due to the fact that the prior approximation, especially for the initial covariance, is not accurate enough in this example. Currently, the covariance is computed by the ERTS in the sliding window where the Jacobian is evaluated at the W-ChevOpt estimated state (note that the state estimates over the current window's time interval are first obtained and then used for covariance propagation). Although this Jacobian evaluation strategy could achieve more accurate covariance than the traditional ERTS, the inconsistency between covariance and state may still occur. Fortunately, the negative effect of prior approximation on the optimization decreases with the increasing of window size and the W-ChevOpt with a bigger window size turns to be more consistent.

To examine the influence of the measurement noise, the measurement covariance in this simulation is reduced to $R_k = 10^2$. The averaged absolute position errors of 100 Monte Carlo runs are plotted in Fig. 13. Comparing with Fig. 6, the position estimation accuracy of all algorithms improves when the measurement covariance decreases. It can be seen that the proposed algorithms are still more accurate than other estimators, except that the batch/fixed-lag ERTS smoother becomes better than the W-ChevOpt in this case. This is because that although the window sizes of the fixed-lag ERTS and W-ChevOpt are identical (3s), the W-ChevOpt has larger window sliding step than the fixed-lag ERTS (3s vs. 1s). At the end time of current sliding window, for example, the W-ChevOpt utilizes the measurements only up to the current time while the fixed-lag ERTS needs future measurements in the forthcoming 3 seconds.

The average time cost across 100 Monte Carlo runs is plotted in Fig. 14. The W-ChevOpt (3s) is more efficient than the fixed-lag ERTS for the sake of a larger window sliding step. The time costs of the W-ChevOpt with different window sizes are comparable to that of the UKF and about 5-6 times larger than that of the EKF. An exception is for the window size of one second. The underlying reason is that a smaller window size leads to a decreased order of the Chebyshev polynomial but an increased number of sliding windows within a given time interval. In this regard, the size of sliding window should be carefully chosen for a specific application.

## VI. DISCUSSIONS AND CONCLUSIONS

This paper proposes a new continuous-time MAP estimation framework, in both batch and sliding window forms. Specifically, the continuous state over the time interval of interest is approximated by the Chebyshev polynomial and then the continuous MAP state estimation is transformed into a problem of Chebyshev coefficient optimization by the way of the Chebyshev collocation method. The proposed batch form, ChevOpt, is an optimal continuous-time estimation in the least squares sense; the sliding window form, the W-ChevOpt, is suboptimal with the Gaussian approximation of the prior distribution. Numerical results of representative examples show that the proposed method results in much improved accuracy over the EKF, UKF and ERTS smoother. A future work will explore more accurate prior approximation, e.g., representing the prior distribution by a Gaussian sum.

APPENDIX [43]

This appendix provides the strategy to handle the non-positiveness of $\mathbf{GQG}^T$ in (5). Without the loss of generality, $\mathbf{G}(t)\mathbf{w}(t)$ in (1) can be transformed to

$$\mathbf{G}(t)\mathbf{w}(t) = \bar{\mathbf{G}}(t)\bar{\mathbf{w}}(t) = \begin{bmatrix} \bar{\mathbf{G}}^{(1)}(t) \\ \bar{\mathbf{G}}^{(2)}(t) \end{bmatrix} \bar{\mathbf{w}}(t) \quad (46)$$

where $\bar{\mathbf{G}}_1(t) \in \mathbb{R}^{r \times r}$ is a nonsingular matrix and $\bar{\mathbf{w}} \in \mathbb{R}^r$ is the Gaussian noise with a positive definite covariance matrix $\bar{\mathbf{Q}}(t)$. Then the dynamics in (1) is rewritten as

$$\dot{\mathbf{x}}(t) = \mathbf{f}(\mathbf{x}(t), \mathbf{u}(t)) + \bar{\mathbf{G}}(t)\bar{\mathbf{w}}(t) \quad (47)$$

Partitioning (47) into two sub-dynamics with $r$ and $n-r$ rows, it yields

$$\begin{bmatrix} \dot{\mathbf{x}}^{(1)}(t) \\ \dot{\mathbf{x}}^{(2)}(t) \end{bmatrix} = \begin{bmatrix} \mathbf{f}^{(1)}(\mathbf{x}(t), \mathbf{u}(t)) \\ \mathbf{f}^{(2)}(\mathbf{x}(t), \mathbf{u}(t)) \end{bmatrix} + \begin{bmatrix} \bar{\mathbf{G}}^{(1)}(t) \\ \bar{\mathbf{G}}^{(2)}(t) \end{bmatrix} \bar{\mathbf{w}}(t) \quad (48)$$

Define a nonsingular matrix $\mathbf{L}(t)$ as

$$\mathbf{L}(t) = \begin{bmatrix} \mathbf{I}_r & \mathbf{0}_{r \times (n-r)} \\ -\bar{\mathbf{G}}^{(2)}(t)(\bar{\mathbf{G}}^{(1)}(t))^{-1} & \mathbf{I}_{n-r} \end{bmatrix} \quad (49)$$

where $\mathbf{0}_{r \times (n-r)}$ denotes a $r \times (n-r)$ zero matrix, and $\mathbf{I}_r$ and $\mathbf{I}_{n-r}$ denote identity matrices of $r$ and $n-r$ dimensions, respectively. Multiplying $\mathbf{L}$ on both sides of (48) yields

$$\dot{\mathbf{y}}(t) = \begin{bmatrix} \dot{\mathbf{y}}^{(1)}(t) \\ \dot{\mathbf{y}}^{(2)}(t) \end{bmatrix} = \begin{bmatrix} \mathbf{g}^{(1)}(\mathbf{y}(t), \mathbf{u}(t)) \\ \mathbf{g}^{(2)}(\mathbf{y}(t), \mathbf{u}(t)) \end{bmatrix} + \begin{bmatrix} \bar{\mathbf{G}}^{(1)}(t)\bar{\mathbf{w}}(t) \\ \mathbf{0}_{(n-r) \times 1} \end{bmatrix} \quad (50)$$

where $\dot{\mathbf{y}}(t) \triangleq \mathbf{L}(t)\dot{\mathbf{x}}(t)$ and the dynamics function $\mathbf{g}(\mathbf{y}(t), \mathbf{u}(t)) \overset{\mathbf{y} \leftarrow \mathbf{x}}{=} \mathbf{L}(t)\mathbf{f}(\mathbf{x}(t), \mathbf{u}(t))$. With the dynamics reformulated in (50), the original state estimation is transformed into that of the new state $\mathbf{y}(t)$, which consists of a sub-dynamics with positive definite noise matrix and a noise-free sub-dynamics.

ACKNOWLEDGEMENT

The authors would like to thank the coordinating editor and anonymous reviewers for their insightful comments and suggestions.